\documentclass[journal]{IEEEtran}
\usepackage{cite}
\ifCLASSINFOpdf
\usepackage[pdftex]{graphicx}
  \graphicspath{{../pdf/}{../jpeg/}}
  \DeclareGraphicsExtensions{.pdf,.jpeg,.png}
\else
  \usepackage[dvips]{graphicx}
  \graphicspath{{../eps/}}
  \DeclareGraphicsExtensions{.eps}
\fi
\usepackage{amsmath}
\usepackage{cases}
\usepackage{amsfonts}
\usepackage{subeqnarray}
\usepackage{subfigure}
\usepackage[ruled,linesnumbered]{algorithm2e}
\makeatletter
\newcommand{\nosemic}{\renewcommand{\@endalgocfline}{\relax}}% Drop semi-colon ;
\newcommand{\dosemic}{\renewcommand{\@endalgocfline}{\algocf@endline}}% Reinstate semi-colon ;
\let\oldnl\nl% Store \nl in \oldnl
\newcommand{\nonl}{\renewcommand{\nl}{\let\nl\oldnl}}% Remove line number for one line
\makeatother
\usepackage[export]{adjustbox} 
\usepackage{booktabs}
\usepackage{setspace}
\usepackage{xcolor}
\usepackage{mathrsfs}
\usepackage{amsmath}
\usepackage{array}
\usepackage{amssymb}
\usepackage{amsthm}

\allowdisplaybreaks
\begin{document}
\setstretch{0.9}
\title{Privacy-Preserving Distributed Parameter Estimation for Probability Distribution of Wind Power Forecast Error}
\author{Mengshuo~Jia, 
		Shaowei~Huang,
		Zhiwen~Wang
and Chen~Shen
%\thanks{This work was supported in part by the Joint Funds of the National Natural Science Foundation of China under Grant U1766206.}
\thanks{M. Jia, S. Huang and C. Shen are with the State Key Laboratory of Power Systems, Tsinghua University, 100084 Beijing, China. Z. Wang is with the Engineering and Efficiency, Bytedance, 100084 Beijing, China.}} % (Emails: jms16@mails.tsinghua.edu.cn, shenchen@mail.tsinghua.edu.cn).}
%\thanks{Y. Wang and G. Hug are with the Power Systems Laboratory, ETH Zurich, 8092 Zurich, Switzerland (Emails: yiwang@eeh.ee.ethz.ch, hug@eeh.ee.ethz.ch).}}
\maketitle

\begin{abstract}
Building the conditional probability distribution of wind power forecast errors benefits both wind farms (WFs) and independent system operators (ISOs). Establishing the joint probability distribution of wind power and the corresponding forecast data of spatially correlated WFs is the foundation for deriving the conditional probability distribution. Traditional parameter estimation methods for probability distributions require the collection of historical data of all WFs. However, in the context of multi-regional interconnected grids, neither regional ISOs nor WFs can collect the raw data of WFs in other regions due to privacy or competition considerations. Therefore, based on the Gaussian mixture model, this paper first proposes a privacy-preserving distributed expectation-maximization algorithm to estimate the parameters of the joint probability distribution. This algorithm consists of two original methods: (1) a privacy-preserving distributed summation algorithm and (2) a privacy-preserving distributed inner product algorithm. Then, we derive each WF's conditional probability distribution of forecast error from the joint one. By the proposed algorithms, WFs only need local calculations and privacy-preserving neighboring communications to achieve the whole parameter estimation. These algorithms are verified using the wind integration data set published by the NREL.
\end{abstract}

\IEEEpeerreviewmaketitle

%\vspace{0.3cm}
%\textbf{\textit{Nomenclature}} 
%\vspace{-0.1cm}
%\begin{table}[h]
%	%\caption{Abbreviation}
%	\label{Abbreviation}
%	\centering
%	\setlength{\tabcolsep}{1mm}{
%	\begin{tabular}{p{2.5cm} l}
%	\toprule
% 	\bfseries $N$ & Total number of observations \\
%	\bfseries $N_i$ & Number of observations in retailer $i$ \\
%	\bfseries $K$ & Total number of clusters \\
%	\bfseries $\boldsymbol{y}_{i,n}$ & The $n$-th observation of retailer $i$ \\
%	\bfseries $\boldsymbol{Y}$ & The union data set of all retailers \\
%	\bfseries $\boldsymbol{\mu}_k$ & The centroid of cluster $k$ \\
%	\bfseries $\boldsymbol{C}_k$ & The set of cluster $k$ \\
%	\bfseries $I(a=b)$ & $I(\cdot)$ equals $1$ if $a=b$ and $0$ otherwise\\
%	\bfseries $m$ & Fuzziness index\\
%	\bfseries $\rho_{k,i,n}^m$ & The degree that  $\boldsymbol{y}_{i,n}$ belongs to $\boldsymbol{C}_k$\\
%	\bfseries $\mathcal N_k(\cdot)$ & The $k$-th Gaussian component of GMM\\
%	\bfseries $\omega_k$ & The weight of $\mathcal N_k(\cdot)$\\
%	\bfseries $\boldsymbol{\Sigma}_k$ & The covariance of $\mathcal N_k(\cdot)$\\
%	\bfseries $\boldsymbol{\varepsilon}$ & The edge set of the graph\\
%	\bfseries $\boldsymbol{\mathcal V}$ & The vertex set of the graph\\
%	\bfseries $\boldsymbol{\Omega}_i$ & The neighborhood of retailer $i$ \\ 
%	\bfseries $d_i$ & The degree of retailer $i$ \\
%	\toprule
%	\end{tabular}}
%\end{table}
%\vspace{-0.5cm}

\section{Introduction}

\IEEEPARstart{A}{s} the penetration of wind power continues to increase in multi-regional interconnected grids \cite{WANG2018945}, a better understanding of wind power forecast error is highly desirable \cite{4530750}. Building the probability distribution of wind power forecast error can benefit both wind farms (WFs) and regional independent system operators (ISOs). For the former, WFs can perform better in market bidding by quantifying stochastic features of their forecast errors \cite{1490597}. For the latter, regional ISOs can make optimal decisions regarding stochastic economic dispatch \cite{7862254} or can schedule enough reserves to meet the demand \cite{7384775}. Note that, if we consider the correlation between WFs when estimating the probability distribution's parameters, the reserve cost will be significantly reduced since the distributions of forecast errors will be more precise \cite{8315454}.

To take the wind power correlation into account, one should first establish the joint probability distribution of the wind power and the corresponding forecast data from correlated WFs. Then, one can derive the conditional probability distribution of the forecast error under a given forecast value from the joint one \cite{wang2018,8481551}. An accurate distribution model is the prerequisite for the above. Therefore, we choose the Gaussian mixture model (GMM) as the distribution model \cite{wang2018}, as the GMM can characterize multivariate random variables subject to an arbitrary distribution with remarkable performance \cite{8481551}.

The expectation-maximization (EM) algorithm is the most commonly used method to estimate the parameters of GMM \cite{SUN2019113842,ZHANG2019229}. The centralized EM algorithm requires collecting all the historical wind power and forecast data to train the GMM-based joint probability distribution. After that, the conditional probability distribution of forecast error under a given forecast value can be derived from it. However, the centralized EM algorithm might not be practical. For example, in a multi-regional interconnected grid, the whole grid is actually controlled by multiple regional ISOs, respectively \cite{982193}. A central operator with access to all data of the whole grid may not exist for political and technical reasons \cite{6939252}. Moreover, regional ISOs are unable to collect the raw data from other areas due to privacy considerations \cite{7990367}. Furthermore, WFs with different stakeholders will not share their forecast values with others, as this may leak their commercial secrets in market bidding \cite{8481551}. Except for the problem of the possible data barriers, the centralized architecture also faces the risk of single-point failures \cite{RAHMAN201720} and the requirement for high bandwidth \cite{chen2019distributed}. To deal with the aforementioned dilemmas, a privacy-preserving distributed (PPD) EM algorithm is a better choice. Specifically, `distributed' means that each WF only needs local calculations and neighboring communications with surrounding WFs, and ``privacy-preserving'' means that the raw data of a WF cannot be deduced by others during the whole estimation process. 

In the data-mining field, many efforts have been made on PPD EM algorithms \cite{Clifton2002,Lin2005,8029559,Yang2012}. The authors of \cite{Clifton2002} and \cite{Lin2005} have proposed a PPD EM algorithm based on the secure sum technique. The algorithm can accurately calculate the sum of data without revealing the data privacy of any parties. A cyclic communication topology is adopted to perform the algorithm. Using the additive homomorphic encryption technique, \textit{Kaleb et al.} present their PPD EM algorithm by encoding the raw data into cryptographic messages \cite{8029559}. To prevent the leakage of data privacy when an adversary controls multiple parties, \textit{Kaleb et al.} enforce one-way communication by a ring topology to guarantee the corruption-resistant feature of the proposed algorithm. Similar to \cite{8029559}, \textit{Yang et al.} also utilize the additive homomorphic encryption technique to protect the raw data \cite{Yang2012}. The differences lie in that (1) \textit{Yang et al.} design a local-global secure summation protocol, and (2) the cryptographic messages are sent through a spanning tree communication topology in \cite{Yang2012}. 

However, the above PPD EM algorithms mentioned in \cite{Clifton2002,Lin2005,8029559,Yang2012}, even including the privacy-free distributed EM algorithms in \cite{5410606,Weng2011,5606758,4558075}, cannot be applied to the joint probability distribution estimation of correlated WFs because all these algorithms are designed for ``horizontally partitioned data''. In fact, wind power and its forecast data are ``vertically partitioned'' among correlated WFs. Take three parties and 100 samples of 3-dimensional random variables as an example. The data being horizontally partitioned refer to the situation where party A owns 30 samples of 3-dimensional data, party B owns 40 samples, and party C owns 30 samples, while the data being vertically partitioned means that all parties own 100 samples but each of them only has one dimension of the 3-dimensional data. Since each WF has only its own historical wind power and forecast data, i.e., two dimensions of the full multidimensional data, wind power and forecast data are actually vertically partitioned among correlated WFs.

Moreover, the PPD EM algorithms in \cite{Clifton2002,Lin2005,8029559,Yang2012} are not fully distributed. Both the preselected cyclic communication topology in \cite{Clifton2002,Lin2005,8029559} and the preselected spanning-tree communication topology in \cite{Yang2012} need a global perspective for preselection. Besides, the failure of any communication line on the preselected path will make the whole algorithm fail. 

In this paper, we aim to solve the above problems and develop a PPD EM algorithm to deal with vertically partitioned wind power and forecast data. This algorithm should enable each WF to obtain the GMM-based joint probability distribution in a fully distributed and privacy-preserving manner, and should be robust to communication line failures. After that, using the PPD derivation algorithm proposed in \cite{Jia2019}, each WF can eventually achieve its conditional probability distribution of forecast error considering the correlation of all WFs. The original contributions of this paper are as follows:

\begin{enumerate}
\item We formulate a distributed framework for the EM algorithm by reformulating the algorithm into local and global parts. Then, the keys to developing a PPD EM algorithm for vertically partitioned wind power and forecast data are pointed out --- to find PPD methods to calculate summations and inner products among the statistics of the correlated WFs.
\item We propose a distributed summation algorithm by leveraging the average consensus algorithm. Then, this algorithm is modified with an additive homomorphic encryption technique to become a PPD summation algorithm. Moreover, we also propose a PPD inner product algorithm on the basis of the randomized binary hash mapping and the average consensus algorithm.
\item Combining the proposed PPD summation and inner product algorithms, we finally propose a PPD EM algorithm to estimate the GMM-based joint probability distribution. This algorithm is fully distributed and strictly protects the raw data of the correlated WFs from leakage. Meanwhile, this algorithm is robust to communication line failures.
\end{enumerate} 

The remainder of this paper is organized as follows. In Section 2, the GMM-based joint and conditional probability distributions are described. In Section 3, the PPD framework for the EM algorithm is formulated to point out the keys to the realization of the PPD EM algorithm. The PPD summation algorithm is developed in Section 4, and the PPD inner product algorithm is designed in Section 5. In Section 6, the PPD EM algorithm is finally proposed. Case studies are provided in Section 7. Section 8 concludes this paper.

\section{GMM-based Probability Distributions}

This section introduces the GMM-based joint and conditional probability distributions. For \textit{M} spatially correlated WFs, their wind power and wind power forecast constitute a \textit{2M}-dimensional random variable $ [\mathbf{X},\mathbf{Y}] \in \mathbb{R}^{2M}$, which is defined as $[\mathbf{X},\mathbf{Y}]=[\boldsymbol{x}_{1},...,\boldsymbol{x}_{M},\boldsymbol{y}_{1},...,\boldsymbol{y}_{M}]$. Elements $\boldsymbol{x}_{m}$ and $\boldsymbol{y}_{m}$ represent the wind power and the forecast of the \textit{m}-th WF, respectively. The GMM-based joint probability distribution function (PDF) of $[\mathbf{X},\mathbf{Y}]$ is actually a convex combination of \textit{J} multivariate Gaussian distributions with its weighted coefficient $w_j \in \mathbb{R}$, mean vector $\boldsymbol{\mu}_j \in \mathbb{R}^{2M}$ and covariance $\boldsymbol{\Sigma}_j \in \mathbb{R}^{2M \times 2M}$, as given in (\ref{Joint GMM}):
\begin{equation}
\label{Joint GMM}
P\big({[\mathbf{X},\mathbf{Y}]}|\boldsymbol{\theta}\big)=\sum\nolimits_{j=1}^J w_j 
\mathcal N\big({[\mathbf{X},\mathbf{Y}]}|\boldsymbol{\mu}_j,\boldsymbol{\Sigma}_j\big)
\end{equation}

\noindent where $\mathcal N(\cdot)$ is a \textit{2M}-dimensional Gaussian distribution and $\boldsymbol{\theta}=\{w_j,\boldsymbol{\mu}_j,\boldsymbol{\Sigma}_j|j=1,2,...,J\}$ is the parameter set of GMM-based joint PDF. The details of $\boldsymbol{\theta}$ are given as follows:
\begin{align}
& \boldsymbol{\mu}_j = 
\begin{bmatrix}
\boldsymbol{\mu}_{j,x} & \boldsymbol{\mu}_{j,y}
\end{bmatrix}, \ \boldsymbol{\mu}_{j,x} =
\begin{bmatrix}
\mu_{j,1} & \cdots & \mu_{j,M} 
\end{bmatrix} \label{mu} \\
& \boldsymbol{\mu}_{j,y} =
\begin{bmatrix}
\mu_{j,M+1} & \cdots & \mu_{j,2M} 
\end{bmatrix}, \ \boldsymbol{\Sigma}_j =
\begin{bmatrix}
\boldsymbol{A}_j & \boldsymbol{B}_j\\
\boldsymbol{B}_j^T & \boldsymbol{C}_j
\end{bmatrix} \label{mu and sigma} \\
& \boldsymbol{B}_j = 
\begin{bmatrix}
\sigma_{j,1,M+1} & \cdots & \sigma_{j,1,2M} \\
\vdots & \vdots & \vdots \\
\sigma_{j,M,M+1} & \cdots & \sigma_{j,M,2M} 
\end{bmatrix} = 
\begin{bmatrix}
\boldsymbol{b}_{j,1} \\
\vdots \\
\boldsymbol{b}_{j,M} 
\end{bmatrix} \label{B} \\
& \boldsymbol{C}_j = 
\begin{bmatrix}
\sigma_{j,M+1,M+1} & \cdots & \sigma_{j,M+1,2M} \\
\vdots & \vdots & \vdots \\
\sigma_{j,2M,M+1} & \cdots & \sigma_{j,2M,2M} 
\end{bmatrix} \label{C} 
\end{align}

Define the forecast error of the $m$-th WF as $\boldsymbol{z}_m = \boldsymbol{x}_m-\boldsymbol{y}_m$. Given a forecast value $\boldsymbol{y}^0 \in \mathbb{R}^{M}$, the conditional PDF of $\boldsymbol{z}_m$ can be derived from (\ref{Joint GMM}) and detailed in (\ref{Conditional GMM}), where its weighted coefficient $\alpha_{j,m}$ is given in (\ref{alpha}), its mean vector $\lambda_{j,m}$ is given in (\ref{lambda}) and its variance $\delta_{j,m}$ is given in (\ref{Delta}). 
\begin{align}
& P(\boldsymbol{z}_m|\boldsymbol{y}^0) = \sum\nolimits_{j=1}^J \alpha_{j,m} 
\mathcal N_j(\boldsymbol{z}_m+\boldsymbol{y}_m|\boldsymbol{y}^0;\lambda_{j,m},\sigma_{j,m}) \label{Conditional GMM} \\
& \alpha_{j,m} = \frac{w_j \mathcal N_j(\boldsymbol{y}^0;\boldsymbol{\mu}_{j,y}, \boldsymbol{C}_j)}{\sum_{j=1}^J w_j 
\mathcal N_j(\boldsymbol{y}^0;\boldsymbol{\mu}_{j,y},\boldsymbol{C}_j)} \label{alpha} \\
& \lambda_{j,m} = \mu_{j,m} + \boldsymbol{b}_{j,m} \boldsymbol{C}_j^{-1} (\boldsymbol{y}^0-\boldsymbol{\mu}_{j,y}) \label{lambda} \\
& \delta_{j,m} = \sigma_{j,m,m} - \boldsymbol{b}_{j,m} \boldsymbol{C}_j^{-1} \boldsymbol{b}_{j,m}^T \label{Delta}
\end{align}

% Once the joint PDF in (\ref{Joint GMM}) is obtained, the conditional PDF in (\ref{Conditional GMM}) can be derived from the joint one. Since the PPD derivation algorithm for conditional PDF is already proposed in \cite{Jia2019}, we mainly focus on developing a PPD EM algorithm to build the joint PDF in this paper.

\section{PPD Framework for the EM Algorithm}

In this section, the centralized EM algorithm for estimating the parameters of GMM is first introduced. Then, we formulate a distributed framework for the EM algorithm by reformulating the algorithm into local and global parts. Specifically, the local parts can be calculated by each WF, and the global parts require the results of all local parts to be computed. To design a PPD EM algorithm, the keys lie in how to achieve the global parts in a PPD manner. Therefore, the distributed framework points out the direction for the PPD EM algorithm design.

\subsection{Centralized EM Algorithm}

The training set $\mathbf{S}\in \mathbb{R}^{N \times 2M}$ consists of \textit{N} historical observations of $[\mathbf{X},\mathbf{Y}]$. The \textit{n}-th observation is described as $\boldsymbol{\varsigma}_n = [s_{n,1},...,s_{n,2M}]$, where $s_{n,m}$ is the \textit{n}-th wind power observation of the \textit{m}-th WF, while $s_{n,M+m}$ is its \textit{n}-th forecast observation. The closed-form expression of the centralized EM algorithm consists of the expectation step (E-step) and maximization step (M-step). In the \textit{k}-th iteration, the centralized E-step is given in (\ref{E-step}) and the centralized M-step in (\ref{M-step}):
\begin{equation}
\label{E-step}
Q_{j,n}^k = \frac{w_j^{k-1} \mathcal N(\boldsymbol{\varsigma}_n;\boldsymbol{\mu}_j^{k-1},\boldsymbol{\Sigma}_j^{k-1})}{\sum_{j=1}^J w_j^{k-1} \mathcal N(\boldsymbol{\varsigma}_n;\boldsymbol{\mu}_j^{k-1},\boldsymbol{\Sigma}_j^{k-1})} 
\end{equation}
\vspace{-10pt}
\begin{subequations}
\label{M-step}
\begin{align}
& w_j^k = \frac{1}{N} \sum_{n=1}^N Q_{j,n}^k \\
& \boldsymbol{\mu}_j^k = \frac{\sum_{n=1}^N Q_{j,n}^k \boldsymbol{\varsigma}_n}{\sum_{n=1}^N Q_{j,n}^k} \\
& \boldsymbol{\Sigma}_j^k = \frac{\sum_{n=1}^N Q_{j,n}^k (\boldsymbol{\varsigma}_n-\boldsymbol{\mu}_j^k)^T(\boldsymbol{\varsigma}_n-\boldsymbol{\mu}_j^k)}{\sum_{n=1}^N Q_{j,n}^k} 
\end{align}
\end{subequations}

\noindent where \textit{T} represents the transpose of a vector or matrix. After convergence, the estimation of $\boldsymbol{\theta}$ is achieved, and GMM-based joint PDF is established. For detailed derivation and proof, please refer to \cite{techrep2}. It should be emphasized that, since the calculation processes are the same for every Gaussian component in every iteration, we will omit the subscripts \textit{k} and \textit{j} in later derivations when it does not cause ambiguity. 

\subsection{Distributed Framework for the E-step}

The E-step aims to calculate the statistics $Q_n$ in (\ref{E-step}) using the parameter $\boldsymbol{\theta}$ updated by each WF from the last iteration. Therefore, in the E-step, $\boldsymbol{\theta}$ becomes public knowledge for all WFs. The part that actually involves $\boldsymbol{\varsigma}_n$ exists only in the exponential term of (\ref{Gaussian distribution}), as given in (\ref{exp term}). 
\begin{align}
& \mathcal N(\boldsymbol{\varsigma}_n|\boldsymbol{\mu}, \boldsymbol{\Sigma}) = 
\frac{exp[-\frac{1}{2} (\boldsymbol{\varsigma}_n-\boldsymbol{\mu}) \boldsymbol{\Sigma}^{-1} (\boldsymbol{\varsigma}_n-\boldsymbol{\mu})^T]}{\sqrt{(2\pi)^{2M} det(\boldsymbol{\Sigma})}} \label{Gaussian distribution} \\
& \epsilon_n 
= (\boldsymbol{\varsigma}_n-\boldsymbol{\mu}) \boldsymbol{\Sigma}^{-1} (\boldsymbol{\varsigma}_n-\boldsymbol{\mu})^T \notag \\
& \ \ \ = \sum\nolimits_{i=1}^{2M} (s_{n,i}-\mu_{i}) \sum\nolimits_{m=1}^{2M} \sigma_{m,i}(s_{n,m}-\mu_{m}) \label{exp term}
\end{align}

Equation (\ref{exp term}) shows that for vertically partitioned data, the E-step can be divided into two summations among WFs: the first one is given in (\ref{sum1}), and the second one is given in (\ref{sum2}). 
\begin{align}
& \tau_{n,i} = \sum\nolimits_{m=1}^{2M} \sigma_{m,i}(s_{n,m}-\mu_{m})=\sum\nolimits_{m=1}^{M}d_{n,i,m}\label{sum1} \\
& \epsilon_n = \sum\nolimits_{i=1}^{2M} \tau_{n,i} (s_{n,i}-\mu_{i})=\sum\nolimits_{m=1}^{M}e_{n,m}\label{sum2}
\end{align}

The local part of the first summation for the $m$-th WF is defined as $d_{n,i,m}$ in (\ref{local1}), where the $m$-th WF can compute it with the known $\boldsymbol{\theta}$ and its own data. The global part of the first summation is (\ref{sum1}) itself. 
\begin{align}
& d_{n,i,m} = \sigma_{m,i}(s_{n,m}-\mu_{m}) \ + \sigma_{M+m,i}(s_{n,M+m}-\mu_{M+m}) \label{local1}
\end{align}

The local part of the second summation for the $m$-th WF is defined as $e_{n,m}$ in (\ref{local2 estep}), while the global part of the second summation is (\ref{sum2}) itself as well.
\begin{align}
& e_{n,m} = \tau_{m}(s_{n,m}-\mu_{m}) \ + \tau_{M+m}(s_{n,M+m}-\mu_{M+m}) \label{local2 estep}
\end{align}

In fact, the relationships between the local and global parts of (\ref{sum1}) and (\ref{sum2}) are the same. Therefore, we provide a unified form of the relationships in (\ref{essence E-step}).
\begin{equation}
\label{essence E-step}
G = \sum\nolimits_{m=1}^M l_m,\ \ \ l_m = d_{n,i,m}\ \ \text{or}\ \ l_m = e_{n,m}
\end{equation}

To achieve the global part $G$, one should collect $l_m$ of others. However, for the \textit{m}-th WF, sharing $l_m$ with others might leak the information of its raw data because the wind power $s_{n,m}$ is close to its forecast value $s_{n,M+m}$; thus, other WFs can estimate the \textit{m}-th WF's data to some extent from $d_{n,i,m}$ or $e_{n,m}$. Therefore, how to calculate (\ref{essence E-step}) in a distributed manner under the premise of data privacy preservation is the key to realizing the PPD E-step.

\subsection{Distributed Framework for the M-step}

The M-step aims to update $\boldsymbol{\theta}$ in (\ref{M-step}) using $Q_n$ calculated from the E-step. For (\ref{M-step}a), since $Q_n$ is already obtained by all WFs as public knowledge, every WF can directly compute (\ref{M-step}a) to update $\omega$. For (\ref{M-step}b), its \textit{m}-th element is reformulated in (\ref{expand mu 2}). The \textit{m}-th WF can compute $\mu_m$ and $\mu_{M+m}$ by itself. Meanwhile, since no WF can deduce \textit{N} observations from the result of (\ref{expand mu 2}), the \textit{m}-th WF can share its $\mu_m$ and $\mu_{M+m}$ with other WFs as no data privacy is sacrificed. Finally, each WF can update $\boldsymbol{\mu}$ using the results of (\ref{expand mu 2}) from others.
\begin{align}
& \mu_m = \sum\nolimits_{n=1}^N Q_n s_{n,m} \bigg/ \sum\nolimits_{n=1}^N Q_n = \sum\nolimits_{n=1}^N c_n s_{n,m} \label{expand mu 2} \\
& c_n = Q_n \bigg/ \sum\nolimits_{n=1}^N Q_n \notag
\end{align}

For (\ref{M-step}c), the diagonal and nondiagonal elements of $\boldsymbol{\Sigma}$ are reformulated in (\ref{diagonal}) and (\ref{non-diagonal}), respectively. The \textit{m}-th WF can directly compute $\sigma_{m,m}$ and $\sigma_{M+m,M+m}$ in (\ref{diagonal}). Besides, neither $\sigma_{m,m}$ nor $\sigma_{M+m,M+m}$ contain private information because no WF can deduce raw data from them. Thus, the \textit{m}-th WF can share $\sigma_{m,m}$ and $\sigma_{M+m,M+m}$ with others. 
\begin{align}
\sigma_{m,m} &  = \sum\nolimits_{n=1}^N Q_{n}(s_{n,m}-\mu_m)^2 \bigg/ \sum\nolimits_{n=1}^N Q_n \notag  \\ 
& = \sum\nolimits_{n=1}^N c_{n}(s_{n,m}-\mu_m)^2 \label{diagonal} \\
\sigma_{m,i} & = \sum\nolimits_{n=1}^N Q_{n}(s_{n,m}-\mu_m) (s_{n,i}-\mu_i) \bigg/ \sum\nolimits_{n=1}^N Q_n \notag  \\ 
& = \sum\nolimits_{n=1}^N c_{n}(s_{n,m}-\mu_m) (s_{n,i}-\mu_i) \label{non-diagonal} \\
& = \left< \boldsymbol{p}_m, \boldsymbol{p}_i \right>, \ m,i = 1,2,...,2M \label{essence M-step} \\
\boldsymbol{p}_m &  = \left [ \sqrt{c_1}(s_{1,m}-\mu_m) \ \
\cdots \ \ 
\sqrt{c_n}(s_{n,m}-\mu_m) \right ] \label{local2} \\
\boldsymbol{p}_i & = \left [ \sqrt{c_1}(s_{1,i}-\mu_i) \ \ \ \ \,
\cdots \ \ \,
\sqrt{c_n}(s_{N,i}-\mu_i) \right ]\label{local3} 
\end{align}

However, the situation is completely different when calculating $\sigma_{m,i}$ in (\ref{non-diagonal}), which requires computing an inner product between vector $\boldsymbol{p}_m \in \mathbb{R}^N$ and $\boldsymbol{p}_i \in \mathbb{R}^N$ in (\ref{essence M-step}). The local part of the inner product is defined in (\ref{local2}) for the $m$-th WF and (\ref{local3}) for the $i$-th WF, where the $m$-th or $i$-th WF can directly compute $\boldsymbol{p}_m$ or $\boldsymbol{p}_i$ with its own data. The global part of the inner product is (\ref{essence M-step}). As can be observed, to obtain the global part, one should collect the local part, i.e., the vector $\boldsymbol{p}_m$ of all WFs. However, since $c_n$ and $\mu_m$ are all public knowledge after the calculations of (\ref{E-step}) and (\ref{M-step}b), collecting $\boldsymbol{p}_m$ is essentially collecting the raw data $s_{n,m}$ of the $m$-th WF, which reveals privacy. Therefore, how to calculate (\ref{essence M-step}) for any two WFs in a distributed manner under the protection of data privacy is the key for realizing the PPD M-step.

\section{PPD Summation Algorithm}

This section proposes a PPD summation algorithm to calculate (\ref{essence E-step}) in a fully distributed and privacy-preserving manner.

\subsection{Average Consensus Algorithm}

To calculate (\ref{essence E-step}) in a fully distributed manner, the average consensus algorithm is an effective approach \cite{4118472}. Some definitions are presented before the demonstration. The communication topology of \textit{M} spatially correlated WFs is represented by a graph $(\boldsymbol{\nu} ,\boldsymbol{\xi})$, where $\boldsymbol{\nu}$ denotes the set of nodes $\boldsymbol{\nu}=\{\nu_1,\nu_2,...,\nu_M\}$ and $\boldsymbol{\xi}$ denotes the set of edges $\boldsymbol{\xi} \subseteq \nu \times \nu$. Once the distance between two nodes is less than a preset distance threshold $\eta$, the two nodes are connected. The neighbors of node \textit{m} are denoted by $\boldsymbol{N}_m=\{\nu_i \in \boldsymbol{\nu}: (\nu_m,\nu_i) \in \xi \}$. The weighted adjacency matrix is represented by $\boldsymbol{A} \in \mathbb{R}^{M \times M}$ with adjacent coefficient $\{ \alpha_{m,i}|m,i=1,...,M\}$ as given in (\ref{adjacent matrix}), where $\left|\boldsymbol{N}_m\right|$ and $\left|\boldsymbol{N}_i\right|$ denote the degree of nodes \textit{m} and \textit{i}. $\boldsymbol{A}$ is a symmetric matrix, and $\boldsymbol{A}\boldsymbol{1}=\boldsymbol{1}$, where $\boldsymbol{1} = \left[1,...,1\right]^T \in \mathbb{R}^{M}$.
\begin{align}
 	\label{adjacent matrix}
 	& \alpha_{m,i}=
		\begin{cases} 
		\frac{1}{\max  ( \left| \boldsymbol{N}_m \right|, \left| \boldsymbol{N}_i \right|)+1} & \nu_i \in \boldsymbol{N}_m,\\
		1-\sum\limits_{\nu_i \in \boldsymbol{N}_m}\frac{1}{\max  ( \left| \boldsymbol{N}_m \right|, \left| \boldsymbol{N}_i \right|)+1} & \nu_i = \nu_i,\\
		0 & \text{others}
		\end{cases}
 \end{align}

The discrete form of the average consensus algorithm is presented in (\ref{neighbor summation}). After convergence, each WF can obtain the average value of (\ref{essence E-step}) in a distributed manner, as given in (\ref{average consensus}). 
\begin{align}
 	G_{m}^{t+1} & = G_m^t + \sum\nolimits_{\nu_i \in N_m} \alpha_{m,i}\left[G_i^t - G_m^t\right] \notag \\ 
 	& = \alpha_{m,m} G_m^{t} + \sum\nolimits_{\nu_i \in N_m} \alpha_{m,i} G_i^t \label{neighbor summation} \\
	G_m^0 & = l_m ,\ \ \lim_{t \to \infty} G_m^t = \frac{1}{M}\sum_{m=1}^M G_m^0 = \frac{1}{M}\sum_{m=1}^M l_m \label{average consensus}	
 \end{align}

To compute (\ref{neighbor summation}), the $m$-th WF only needs to collect $\alpha_{m,i} G_i^t$ (for $\nu_i \in N_m$) from its neighbors to calculate a local summation during each iteration, i.e.,
\begin{align}
	\Xi_m^t = \sum\nolimits_{\nu_i \in N_m} \alpha_{m,i} G_i^t \label{local summation}
\end{align}

However, in the first iteration, $G_i^0=l_i$ (for $\nu_i \in N_m$) is revealed to the $m$-th WF. Thus, the average consensus algorithm is not privacy-preserving. 

\subsection{PPD Summation Algorithm}

To achieve the local summation in (\ref{local summation}) under the premise of protecting privacy, we leverage an additive homomorphic encryption technique named Paillier cryptosystem. The Paillier cryptosystem is favored by many researchers for the analysis of social networks \cite{Yang2012} or the Internet \cite{5601452}. 

Let $\boldsymbol{pt} \in \mathbb{Z} $ denote a plaintext, $\boldsymbol{ct} \in \mathbb{Z} $ denote a ciphertext and $H$ denote a prespecified large prime integer. The encryption process with a public key $pk$ is given in (\ref{encryption}), and the decryption with a secret key $sk$ is given in (\ref{decryption}). 
\begin{align}
& \boldsymbol{ct} \, = E_{pk}(\boldsymbol{pt},H) \label{encryption} \\
& \boldsymbol{pt} = D_{sk}(\boldsymbol{ct},H) \label{decryption} 
\end{align}

To compute the sum of \textit{M} plaintexts, the decrypter only needs to multiply the corresponding \textit{M} ciphertexts and then decrypt the multiplication result, as given in (\ref{additive}). The entire process strictly protects data privacy. See \cite{Paillier1999} for more details.
\begin{align}
& D_{sk} \left ( \prod_{m=1}^M \boldsymbol{ct}_m \ \text{mod} \ H^2 \right ) 
= \sum_{m=1}^M \boldsymbol{pt}_m \ \ \text{mod} \ \ H \label{additive} 
\end{align}

Inspired by the secure summation protocol in \cite{Yang2012}, we utilize the Paillier cryptosystem to compute (\ref{local summation}), which helps us to realize a privacy-preserving average consensus algorithm. Specifically, in the first iteration of the average consensus algorithm, the neighbors of the \textit{m}-th WF, which are numbered from 1 to $|N_m|$, encrypt their initial value using the 1st neighbor's $pk$ in (\ref{NeighborEncrypt}). Meanwhile, the \textit{m}-th WF encrypts a random and secret number $R_0$ by (\ref{MEncrypt}). Then, these neighbors send their ciphertexts to the \textit{m}-th WF. After that, the \textit{m}-th WF performs the multiplication calculation in (\ref{Multiply}) and sends the result to the 1st neighbor. Thereafter, the 1st neighbor decrypts $ct$ into the summation in (\ref{1Decrypt}) and sends it back to the \textit{m}-th WF. Finally, the \textit{m}-th WF subtracts $R_0$ to obtain the result of (\ref{local summation}). For the subsequent iterations of the average consensus algorithm, no encryption is needed. Details of the PPD summation algorithm are given in Algorithm 1.  
\begin{align}
& ct_i = E_{pk}(\alpha_{m,i}G_i^0,H), \ \ \forall \nu_i \in \boldsymbol{N}_m \label{NeighborEncrypt} \\
& ct_m = E_{pk}(R_0,H), \ \ \text{for} \ \nu_m \label{MEncrypt} \\
& ct_d = ct_m \times \prod\nolimits_{i=1}^{|\boldsymbol{N}_m|}ct_i \label{Multiply} \\
& D_{sk}\left ( ct_d \ \text{mod} \ H^2 \ \right) = \sum\nolimits_{i=1}^{|\boldsymbol{N}_m|}\alpha_{m,i}G_i^0 + R_0 \label{1Decrypt}
\end{align} 

Please note that although the Paillier cryptosystem is introduced in the average consensus algorithm, it does not affect the convergence of the average consensus algorithm. In fact, the PPD summation algorithm and the average consensus algorithm are mathematically equivalent. See \cite{XIAO200733} for the convergence proof of the original average consensus algorithm.

\begin{algorithm}
\caption{The PPD summation algorithm}
\KwIn{$\forall \nu_m \in \nu$ with its $l_m$.} 
\KwOut{$\forall \nu_m \in \nu$ obtains $G=\sum_{m=1}^Ml_m$.}
\While{convergence criterion is not met}
{
\For{$m=1$ to $M$}
{
$t=0$\;
\eIf{t=0}
{
$\forall \nu_i \in N_m$ computes (\ref{NeighborEncrypt})\;
$\nu_m$ computes (\ref{MEncrypt}) and sends $ct_m$ to $\nu_1$\;
$\nu_1$ computes (\ref{1Decrypt}) and sends it to $\nu_m$ \;
$\nu_m$ subtracts $R_0$ and obtains $\sum_{i=1}^{|\boldsymbol{N}_m|}\alpha_{m,i}G_i^0$\;
$\nu_m$ completes the calculation of (\ref{local summation})\;
$t=t+1$\;
}
{
$\forall \nu_i \in N_m$ sends $\alpha_{m,i}G_i^t$ to $\nu_m$ \;
$\nu_m$ completes the calculation of (\ref{local summation})\;
$t=t+1$\;
}
}
}
\For{$m=1$ to $M$}
{ $\nu_m$ computes $\sum_{m=1}^Ml_m = M \times G_m$
}
\end{algorithm}
\vspace{-0.3cm}
\section{PPD Inner Product Algorithm}

In this section, a PPD inner product algorithm is proposed to calculate (\ref{essence M-step}) for any two WFs in a fully distributed manner considering privacy protection.

For the inner product calculation in (\ref{essence M-step}), once the angle $\beta_{m,i}$ in (\ref{Cauchy}) is obtained, by sharing the norm $\|\boldsymbol{p}_m \| \in \mathbb{R} $ and $\|\boldsymbol{p}_i \| \in \mathbb{R}$ among WFs, which will not reveal any raw data, the inner product can be directly calculated by every WF. Therefore, the problem of computing (\ref{essence M-step}) becomes into this one: how to compute the angle between two vectors in (\ref{Cauchy}) without revealing any raw element of the vectors? 
\begin{align}
& \left< \boldsymbol{p}_m,\boldsymbol{p}_i \right> = \|\boldsymbol{p}_m \| \|\boldsymbol{p}_i \|\cos\beta_{m,i} \label{Cauchy}
\end{align}

In an \textit{N}-dimensional space, the probability of finding a random hyperplane separating two vectors $\boldsymbol{p}_m$ and $\boldsymbol{p}_i$ is proportional to the angle $\beta_{m,i}$ \cite{Goemans1995}. For calculating the probability, a publicly known random vector set $\boldsymbol{\Gamma} \in \mathbb{R}^{N \times L}$ is first defined, where each column is a random vector $\boldsymbol{\gamma}_l \in \mathbb{R}^{N}$. Then, the probability can be computed by (\ref{probability hyperplane}). 
\begin{align}
& Pr\{ \boldsymbol{\gamma}_l \in \Gamma: \left( \boldsymbol{p}_m^T \boldsymbol{\gamma}_l \right) \left( \boldsymbol{p}_i^T \boldsymbol{\gamma}_l \right) < 0\} = \frac{\beta_{m,i}}{\pi} \label{probability hyperplane} 
\end{align}

For a further demonstration, the randomized binary hash mapping function $h\colon \mathbb{R}^{N}\mapsto \mathbb{R}^{L}$ is defined in (\ref{hash}), 
\begin{align}
& h\left(\boldsymbol{p}_m \right) = sign\left(\boldsymbol{p}_m^T \boldsymbol{\Gamma} \right) \label{hash}
\end{align}
where the $sign$ function can encode an \textit{L}-dimensional real vector into an \textit{L}-dimensional binary vector according to the sign of each element in the real vector. Thus, $h\left(\boldsymbol{p}_m\right)$ actually represents the sign information of the multiplication results between $\boldsymbol{p}_m$ and $\forall \boldsymbol{\gamma}_l \in \boldsymbol{\Gamma}$. 

Once $h\left(\boldsymbol{p}_m\right)$ and $h\left(\boldsymbol{p}_i\right)$ are obtained, (\ref{probability hyperplane}) can be easily computed by counting the number of different sign pairs. Note that the counting process is essentially calculating the Hamming distance between $h\left(\boldsymbol{p}_m\right)$ and $h\left(\boldsymbol{p}_i\right)$, i.e., $H_{am}[h\left(\boldsymbol{p}_m\right),h\left(\boldsymbol{p}_i\right)]$ \cite{MARUKATAT2013}. Therefore, based on the randomized binary hash mapping function and Hamming distance calculation, the angle $\beta_{m,i}$ can be calculated by (\ref{angle}). 
\begin{align}
& \beta_{m,i} = \frac{\pi}{L} H_{am}[h\left(\boldsymbol{p}_m\right),h\left(\boldsymbol{p}_i\right)] \label{angle}
\end{align}

It should be emphasized that, our goal is not only to calculate the inner product of two vectors under the premise of protecting privacy but also to obtain all the inner products between any two WFs through a fully distributed manner. For computing all the inner product values, the set $\{\|\boldsymbol{p}_m \|,h\left(\boldsymbol{p}_m\right)|m=1,...,2M\}$ is required. Thus, based on (\ref{angle}) and the average consensus algorithm, the PPD inner product algorithm is proposed in Algorithm 2. 
\begin{algorithm}
\caption{The PPD inner product algorithm}
\KwIn{$\forall \nu_m \in \nu$ with its $\boldsymbol{p}_m$ and $\boldsymbol{p}_{M+m}$.} 
\KwOut{$\forall \nu_m \in \nu$ obtains $\sigma_{m,i} = \left< \boldsymbol{p}_m,\boldsymbol{p}_i \right>$ for $m,i=1,...,2M$}
$\forall \nu_m \in \nu$ computes its $h\left(\boldsymbol{p}_m\right)$ and $h\left(\boldsymbol{p}_{M+m}\right)$ by (\ref{hash}) and converts them into decimal $d_m$ and $d_{M+m}$\;
$\forall \nu_m \in \nu$ formulates its $\boldsymbol{G}_m^0 = [...,d_m,...,d_{M+m},...]$, where $d_m$ is the \textit{m}-th element and $d_{M+m}$ is the (\textit{M+m})-th element. Other elements are 0\;
$\forall \nu_m \in \nu$ computes (\ref{neighbor summation}) with $\boldsymbol{G}_m^0$ as input until convergence \;
$\forall \nu_m \in \nu$ multiplies the convergence result by \textit{M} to obtain $d_m$ for $m=1,...,2M$\;
$\forall \nu_m \in \nu$ converts $d_m$ into binary $h\left(\boldsymbol{p}_m\right)$ for $m=1,...,2M$\;
$\forall \nu_m \in \nu$ computes its $\|\boldsymbol{p}_m \|$ and $\|\boldsymbol{p}_{M+m} \|$\;
$\forall \nu_m \in \nu$ repeats step 2 to 4 by replacing $d_m$ with $\|\boldsymbol{p}_m \|$ and $d_{M+m}$ with $\|\boldsymbol{p}_{M+m} \|$ \;
$\forall \nu_m \in \nu$ computes (\ref{angle}) to obtain $\beta_{m,i}$  ($m,i=1,...,2M$) using $h\left(\boldsymbol{p}_m\right)$ ($m=1,...,2M$)\;
$\forall \nu_m \in \nu$ computes (\ref{Cauchy}) to obtain $\sigma_{m,i}=\left< \boldsymbol{p}_m,\boldsymbol{p}_i \right>$ ($m,i=1,...,2M$) using $\beta_{m,i}$  ($m,i=1,...,2M$) and $\|\boldsymbol{p}_m \|$ ($m=1,...,2M$); 
\end{algorithm}

\section{PPD EM Algorithm}

Since the two keys mentioned in Section 3 are solved by the proposed PPD summation and inner product algorithm, the PPD EM algorithm for estimating the GMM-based joint PDF of multiple spatially correlated WFs is eventually developed in Algorithm 3. 
\begin{algorithm}
\caption{The PPD EM algorithm}
\textbf{Initialization:}\\
01 Set $\omega_j^0$, $\boldsymbol{\mu}_j^0$ and $\boldsymbol{\Sigma}_j^0$ for $j=1,...,J$\;
02 Set k=1\;
\textbf{The PPD E-step:}\\
\For{$j=1$ to $J$ \rm{and} $n=1$ to $N$}
{
03 Define $\omega=\omega_j^{k-1}$, $\boldsymbol{\mu}=\boldsymbol{\mu}_j^{k-1}$, $\boldsymbol{\Sigma}=\boldsymbol{\Sigma}_j^{k-1}$\;
04 Input: $\forall \nu_m \in \nu$ with $l_m = d_{n,i,m}$. Run Algorithm 1 and $\forall \nu_m \in \nu$ obtains $\tau_{n,i}=\sum_{m=1}^{M} d_{n,i,m}$\;
05 Input: $\forall \nu_m \in \nu$ with $l_m = e_{n,m}$. Run Algorithm 1 and $\forall \nu_m \in \nu$ obtains $\epsilon_n= \sum_{m=1}^{M}e_{n,m}$\; 
06 $\forall \nu_m \in \nu$ computes (\ref{Gaussian distribution}) via $\epsilon_n$, and then updates $Q_{j,n}^k$ in (\ref{E-step}) via the result of (\ref{Gaussian distribution})\;
}
\textbf{The PPD M-step:}\\

\For{$j=1$ to $J$}
{
07 $\forall \nu_m \in \nu$ updates $\mu_m$ and $\mu_{M+m}$ in (\ref{expand mu 2})\;
08 $\forall \nu_m \in \nu$ updates $\sigma_{m,m}$ and $\sigma_{M+m,M+m}$ in (\ref{diagonal})\;
09 $\forall \nu_m \in \nu$ updates $\boldsymbol{p}_m$ and $\boldsymbol{p}_{M+m}$ in (\ref{essence M-step})\;
10 $\forall \nu_m \in \nu$ formulates $\boldsymbol{G}_m^0 = [...,\mu_m,...,\mu_{M+m},...]$, where $\mu_m$ is the \textit{m}-th element and $\mu_{M+m}$ is the (\textit{M+m})-th element. Other elements are 0\;
11 $\forall \nu_m \in \nu$ computes (\ref{neighbor summation}) with its $\boldsymbol{G}_m^0$ as input until convergence \;
12 $\forall \nu_m \in \nu$ multiplies the convergence result by \textit{M} to obtain $\mu_m$ for $m=1,...,2M$\; 
13 Repeat step 10 to step 12 while replacing $\mu_m$ and $\mu_{M+m}$ by $\sigma_{m,m}$ and $\sigma_{M+m,M+m}$\; 
14 Input: $\forall \nu_m \in \nu$ with $\boldsymbol{p}_m$ and $\boldsymbol{p}_{M+m}$. Run Algorithm 2 and $\forall \nu_m \in \nu$ obtains $\sigma_{m,i}$ for $m,i=1,...,2M$\;
15 $\forall \nu_m \in \nu$ obtains $\sigma_{m,i}$ for $m,i=1,...,2M$ by (\ref{essence M-step})\;
16 $\forall \nu_m \in \nu$ obtains $\omega_{j}^k$ by (\ref{M-step}) directly\;
17 $\forall \nu_m \in \nu$ updates $\boldsymbol{\mu}_j^k= [\mu_1,...,\mu_{2M}]$\; 
18 $\forall \nu_m \in \nu$ obtains $\boldsymbol{\Sigma}_j^k= \begin{bmatrix}
\sigma_{1,1} & \cdots & \sigma_{1,2M} \\
\vdots & \ddots & \vdots \\
\sigma_{2M,1} & \cdots & \sigma_{2M,2M}
\end{bmatrix}$\;
}
19 Set $k=k+1$\;
20 Loop the PPD E-step and M-step until convergence\;
\end{algorithm}

The advantages of the proposed algorithm are as follows:

\begin{itemize}
	\item \textit{\textbf{Strict privacy-preserving}}. For the summation calculation in the PPD E-step, the Paillier cryptosystem is used to protect the raw data; for the inner product calculation in the PPD M-step, the randomized binary hash mapping function is used to prevent data privacy disclosure. The two techniques that we utilized can strictly protect the privacy of each WF.
	\item \textit{\textbf{Fully distributed}}. As we introduce the average consensus algorithm into the PPD E-step and M-step, each WF only needs to communicate with its neighbors. Thus, we avoid the assumption made in \cite{5601452,Clifton2002,Clifton20022} that any two nodes are connected, and we improve the scalability of the proposed algorithm. Meanwhile, the preselected communication paths in \cite{Clifton2002,Lin2005,8029559,Yang2012} are no longer required. Thus, the proposed algorithm is fully distributed. 
	\item \textit{\textbf{Robust}}. As the communication between neighbors may fail, robustness to communication failure is necessary. Since the proposed PPD EM algorithm is developed based on the average consensus algorithm, as long as the communication topology is still connected, the communication failure basically will not affect the final estimation results due to the consensus feature \cite{4542554}.
\end{itemize}

Please note that once the joint PDF is established, each WF can derive its conditional PDF of the forecast error in (\ref{Conditional GMM}) via the PPD derivation algorithm presented in \cite{Jia2019}.

\vspace{-0.3cm}
\section{Case Study}

The historical data of wind power and forecast value are from the “eastern wind integration data set” published by the National Renewable Energy Laboratory (NREL), the U.S., where we choose 9 WFs in Maryland, numbered 4401, 5405, 6211, 6359, 6526, 6812, 6931, 7187, and 7460. Their communication topology is shown in  Fig. \ref{case_topo_aver}(a). Each WF has 20 days of hourly wind power and forecast data. Thus, $M=9$ and $N=480$. Then we build the joint PDF of the wind power and its forecast of the nine spatially correlated WFs by the proposed EM algorithm. After that, by leveraging the PPD derivation algorithm presented in \cite{Jia2019}, we also derive the conditional PDF of the forecast error of each WF from the joint PDF. Since the privacy-preserving feature of the proposed algorithm has already been discussed in the previous sections, this section mainly aims to verify the correctness and robustness of the proposed algorithm.
\vspace{-0.3cm}
\subsection{Correctness of the PPD Summation Algorithm}

\begin{figure}[h]
\centering  
\subfigure[]{ 
\includegraphics[width=1.3in]{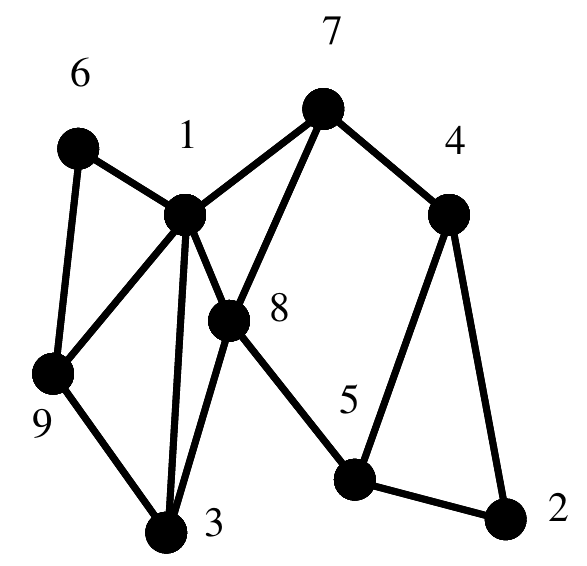} 
} 
\subfigure[]{ 
\includegraphics[width=1.9in]{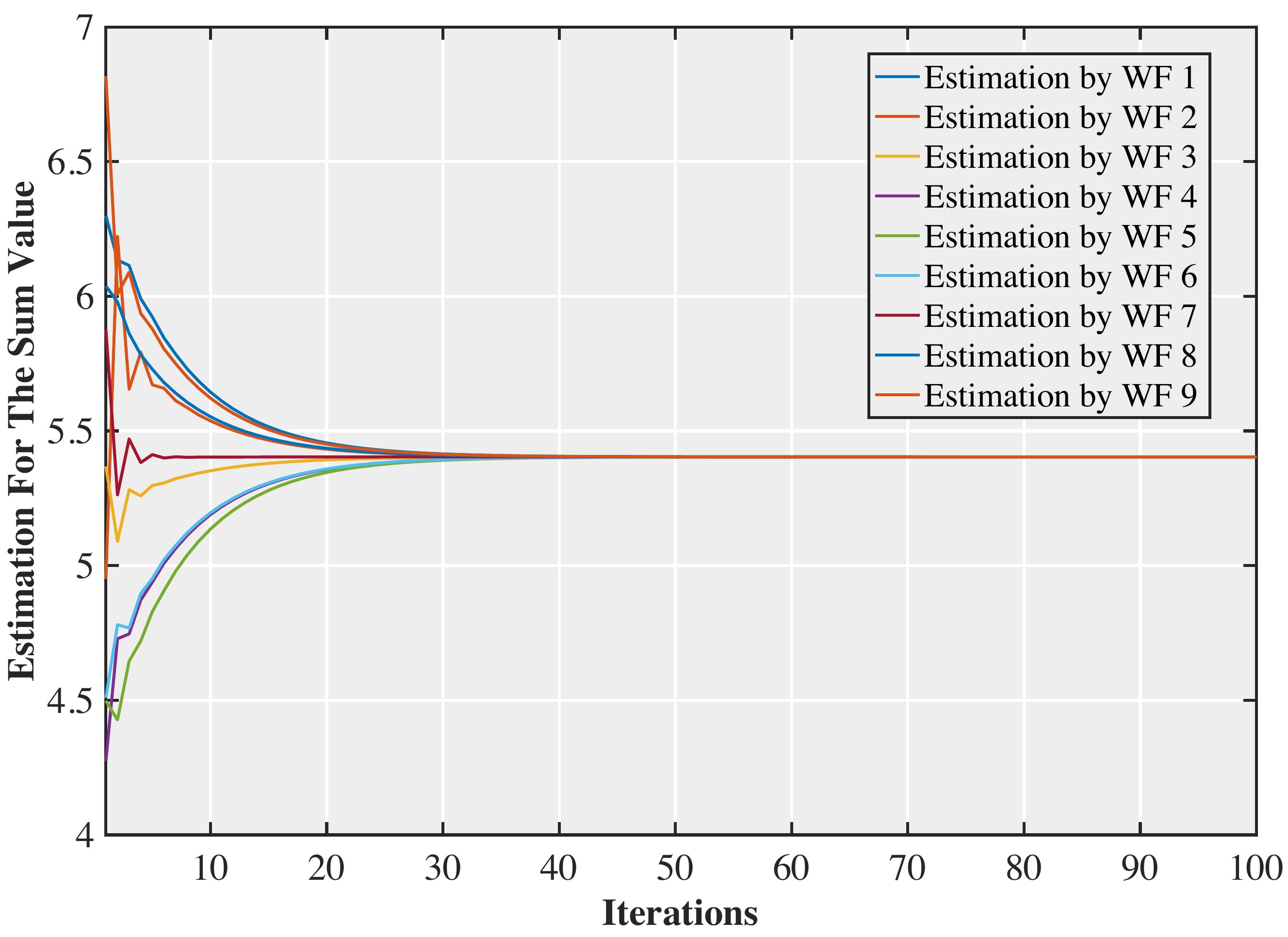} 
} 
\caption{(a) Communication topology of the nine WFs; (b) the iterative process of each WF's estimation for the summation }
\label{case_topo_aver} 
\end{figure}

We use the wind power data of 9 WFs at 2004/1/1-01:00 as input, and use the proposed PPD summation algorithm to estimate the summation of the 9 data points. To show more details, we illustrate the iterative process of each WF's estimation for the summation in Fig. \ref{case_topo_aver}(b). It can be observed that after 30 iterations, all WFs achieve consensus on the exact summation value, showing the correctness of the proposed PPD summation algorithm.
\vspace{-0.5cm}

\vspace{-0.3cm}
\subsection{Correctness of the PPD Inner Product Algorithm}

\begin{figure}[h]
\centering 
\includegraphics[width=2.15in,center]{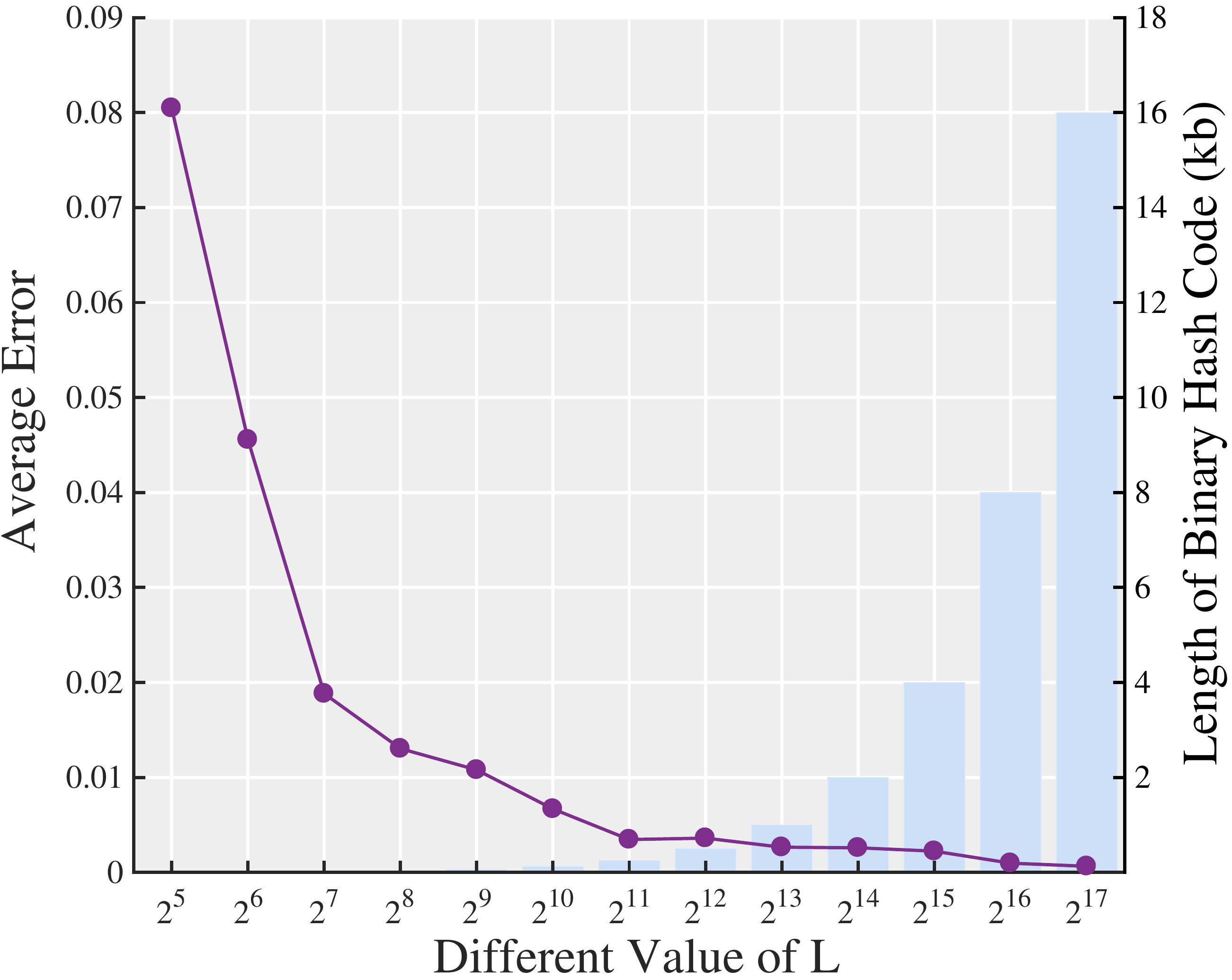}
\caption{Average relative error and its corresponding binary code length}
\label{case_length}
\end{figure}

To verify the correctness of the proposed PPD inner product algorithm, we use it to calculate the inner products between every two WFs' private vectors. The private vector of a WF consists of its 20-day historical wind power data. Then, compared to the real inner product results, we provide the average relative errors of the proposed algorithm under different values of $L$ in Fig. \ref{case_length}. It can be observed that the error decreases significantly as $L$ increases, but it nearly stabilizes when $L$ reaches $2^{11}$. Thus, we finally choose $L$ as $2^{11}$ bit $=0.25$ kb. Furthermore, the average relative error is $3.5\times 10^{-3}$ when $L=2^{11}$, which proves the correctness of the proposed algorithm.
\vspace{-0.3cm}

\vspace{-0.5cm}
\subsection{Verification of the PPD EM Algorithm}

We use the Bayesian information criterion (BIC) to set the number of Gaussian components $J$ as 5. After that, we build the joint probability distribution of wind power and forecast of the nine spatially correlated WFs using the proposed PPD EM algorithm. The distribution constructed by the centralized EM algorithm is also given as the benchmark. Since the 18-dimensional distribution cannot be drawn for illustration, we derive several 1-dimensional and 2-dimensional distributions from the 18-dimensional one based on the linear invariance property of GMM \cite{8481551}. 

\begin{figure}[h]
\setlength{\abovecaptionskip}{0pt}
\centering 
\includegraphics[width=2.5in,center]{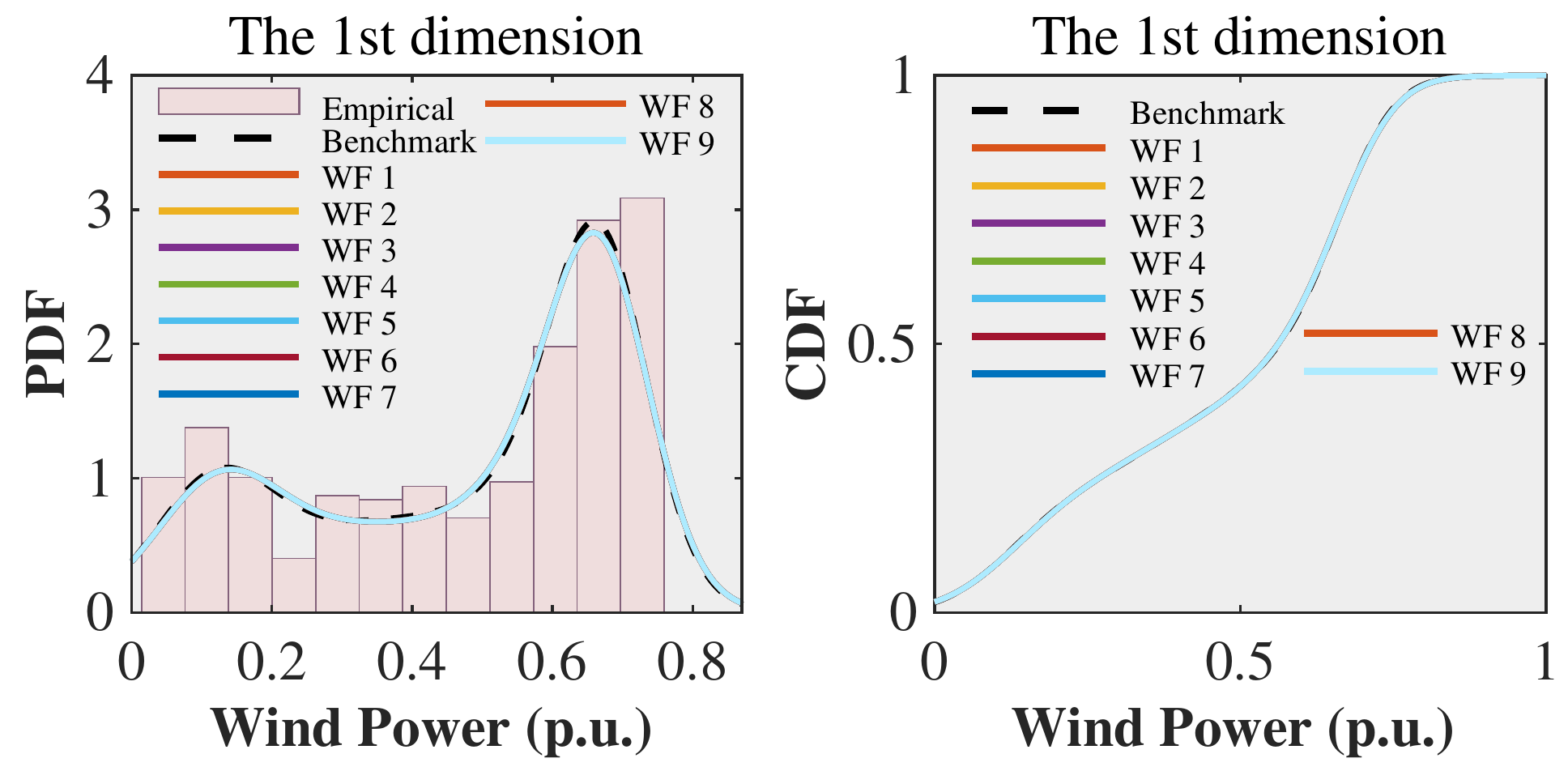}
\caption{The marginal PDFs and CDFs comparisons}
\label{case_PCDF1}
\end{figure}

First, the 1-dimensional PDF and the 1-dimensional cumulative distribution function (CDF) are shown in Fig. \ref{case_PCDF1}. Only the first dimension is provided. 
\vspace{-0.3cm}

In Fig. \ref{case_PCDF1}, the empirical distributions are obtained from the corresponding original historical data, the benchmarks are built by the centralized EM algorithm, and the distribution of each WF is constructed by the proposed PPD EM algorithm. It can be observed that (1) the benchmark and the distributions built by the WFs match the corresponding empirical distributions; (2) the distributions built by the WFs are coincident with each other, indicating that the consensus among WFs is achieved by the proposed algorithm; and (3) the distribution built by each WF is coincident with the benchmark, indicating the correctness of the proposed algorithm.

Then, further comparisons between the benchmark and the distribution of each WF are made using the relative standard error (RSE) as defined in (\ref{RSE}), where $f(\cdot)$ represents the PDF or CDF built by each WF, $f_0(\cdot)$ represents the benchmark PDF or CDF, and $\overline{f_0(\cdot)}$ represents the mean value. The RSE results are provided in Table. \ref{Table_RSE}. First, all the RSE values are less than $2.4\times 10^{-3}$, which means that the difference between the benchmark and the distribution of each WF is tiny. Second, the consensus effect of the proposed algorithm is obvious as the RSE values are identical correspondingly. Third, the RSE values of the CDFs are much smaller than those values of the PDFs. Note that the CDF is what we ultimately want for optimal decisions, e.g., calculating the quantile from the CDF. Thus, the RSE values of the CDFs eventually represent the accuracy of the proposed algorithm.
\begin{equation}
	\label{RSE}
	RSE = \frac{\sum_{n=1}^N\left[f(x_n)-f_0(x_n)\right]^2}{\sum_{n=1}^N\left[\overline{f_0(x_n)}-f_0(x_n)\right]^2}
\end{equation}
\vspace{-0.3cm}
\begin{table}[h]
\renewcommand{\arraystretch}{1.3}
\caption{The RSEs between the distributions built by the centralized EM algorithm and the proposed algorithm}
\label{Table_RSE}
\centering
\footnotesize
\begin{tabular}{p{1.9cm} p{0.3cm} p{0.3cm} p{0.3cm} p{0.3cm} p{0.3cm} p{0.3cm} p{0.3cm} p{0.3cm} p{0.3cm} }
\toprule
\bfseries Wind Farm & $1$ & $2$ & $3$ & $4$ & $5$ & $6$ & $7$ & $8$ & $9$ \\
\hline
\bfseries PDF ($\times 10^{-3}$) & 2.4 & 2.4 & 2.4 & 2.4 & 2.4 & 2.4 & 2.4 & 2.4 & 2.4 \\
\bfseries CDF ($\times 10^{-5}$) & 4.8 & 4.8 & 4.8 & 4.8 & 4.8 & 4.8 & 4.8 & 4.8 & 4.8 \\
\hline
\toprule
\end{tabular}
\end{table}

\begin{figure}[h]
\centering 
\subfigure{ 
\includegraphics[width=2.5in]{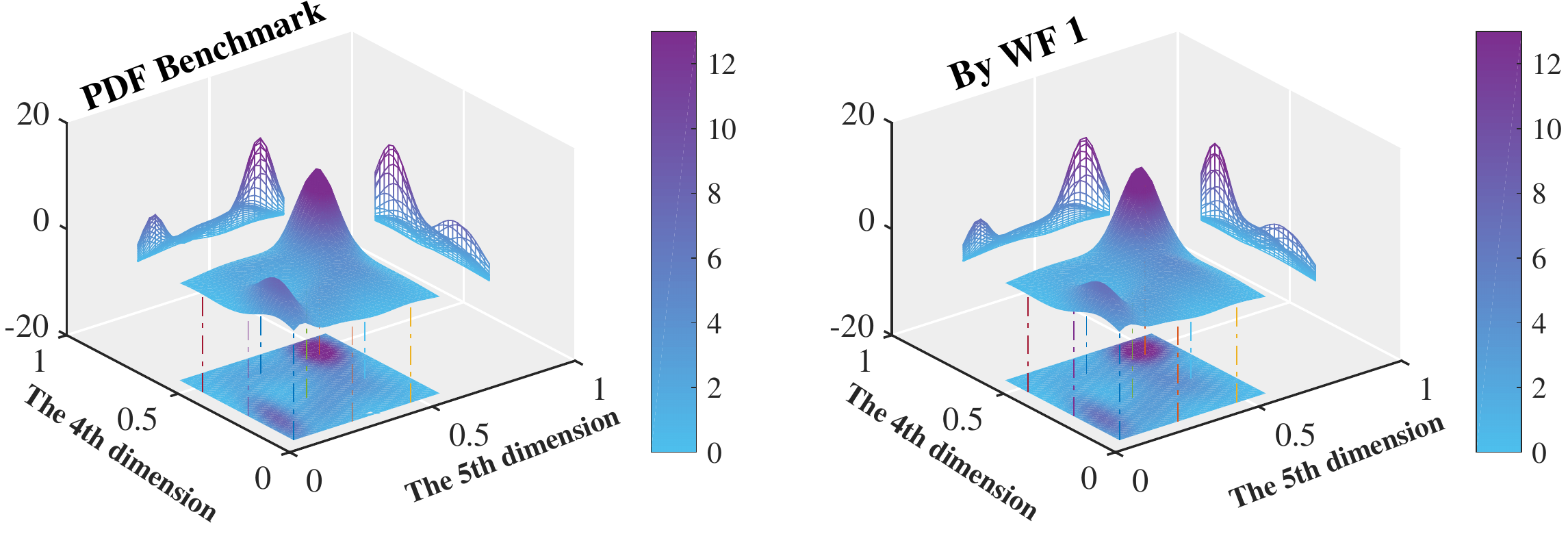} 
} 
\subfigure{ 
\includegraphics[width=2.5in]{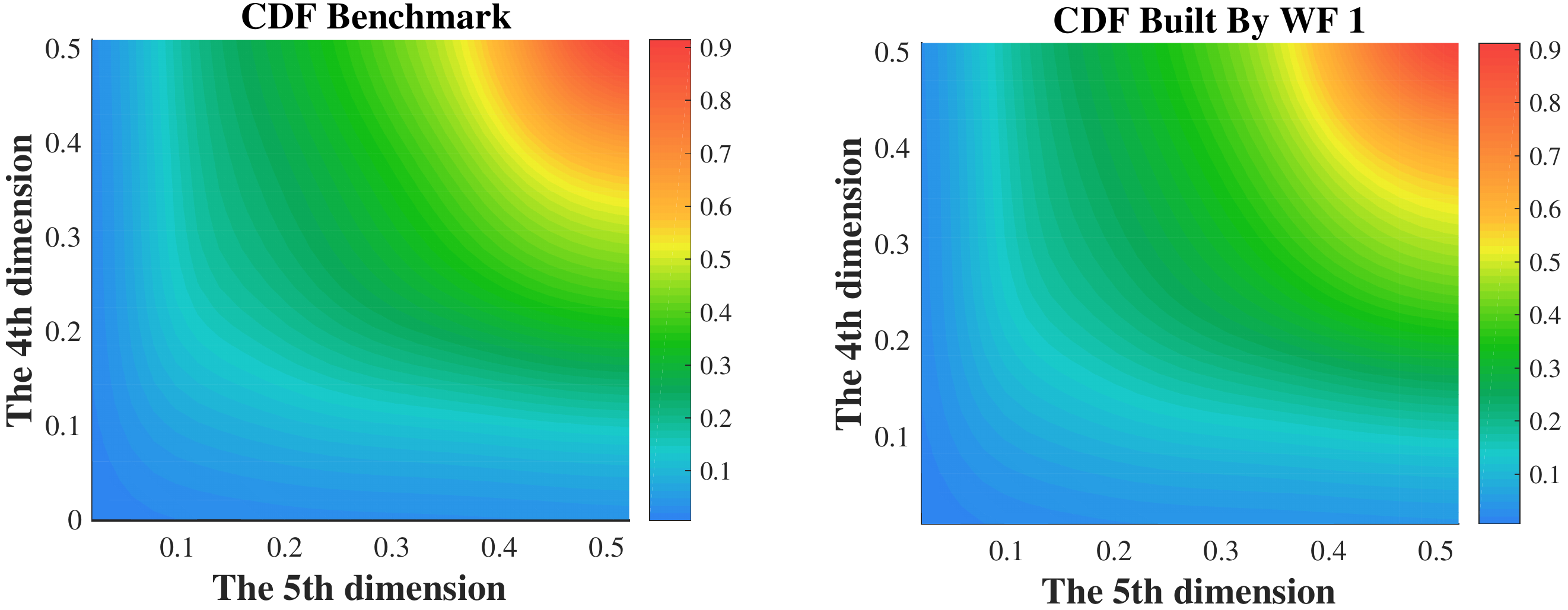} 
} 
\caption{The 2-dimensional PDFs and CDFs comparison}
\label{case_PCDF2} 
\end{figure}

Furthermore, we also choose two dimensions from the 18-dimensional joint distribution to form 2-dimensional PDF and CDF. The 2-dimensional benchmarks built by the centralized EM algorithm and the 2-dimensional joint distributions built by the 1st WF are illustrated in Fig. \ref{case_PCDF2}. 

Thereafter, the Kullback--Leibler divergence (KLD) between the joint distribution built by the 1st WF and by other WFs are given in Table \ref{Table_KL} to illustrate the differences between them. Since all the KLDs are less than $2.19\times 10^{-15}$, the distribution built by the 1st WF and those built by others are basically the same. Thus, using the result of the 1st WF as a representative is reasonable and acceptable. We can observe that, in Fig. \ref{case_PCDF2}, the 2-dimensional PDF and CDF built by the 1st WF match the 2-dimensional benchmarks well. Therefore, the correctness of the proposed PPD EM algorithm is eventually verified.
\vspace{-0.3cm}
\begin{table}[h]
\renewcommand{\arraystretch}{1.3}
\caption{KL divergences between the distributions built by the 1st WF and those built by other WFs}
\label{Table_KL}
\centering
\footnotesize
\begin{tabular}{p{1.9cm} p{0.3cm} p{0.3cm} p{0.3cm} p{0.3cm} p{0.3cm} p{0.3cm} p{0.3cm} p{0.3cm} p{0.3cm} }
\toprule
\bfseries Wind Farm & $1$ & $2$ & $3$ & $4$ & $5$ & $6$ & $7$ & $8$ & $9$ \\
\hline
\bfseries KLD ($\times 10^{-15}$) & 0 & 0.02 & 0.19 & 2.19 & 1.09 & 0.33 & 1.11 & 2.06 & 1.01 \\
\hline
\toprule
\end{tabular}
\end{table}

\vspace{-0.3cm}
\subsection{Robustness of the PPD EM Algorithm}

Since the proposed PPD EM algorithm is developed based on the average consensus algorithm, as long as the communication network topology is still connected, the communication failure basically will not affect the final estimation results due to the consensus feature. 
\vspace{-0.3cm}
\begin{figure}[h]
\centering 
\includegraphics[width=2.5in,center]{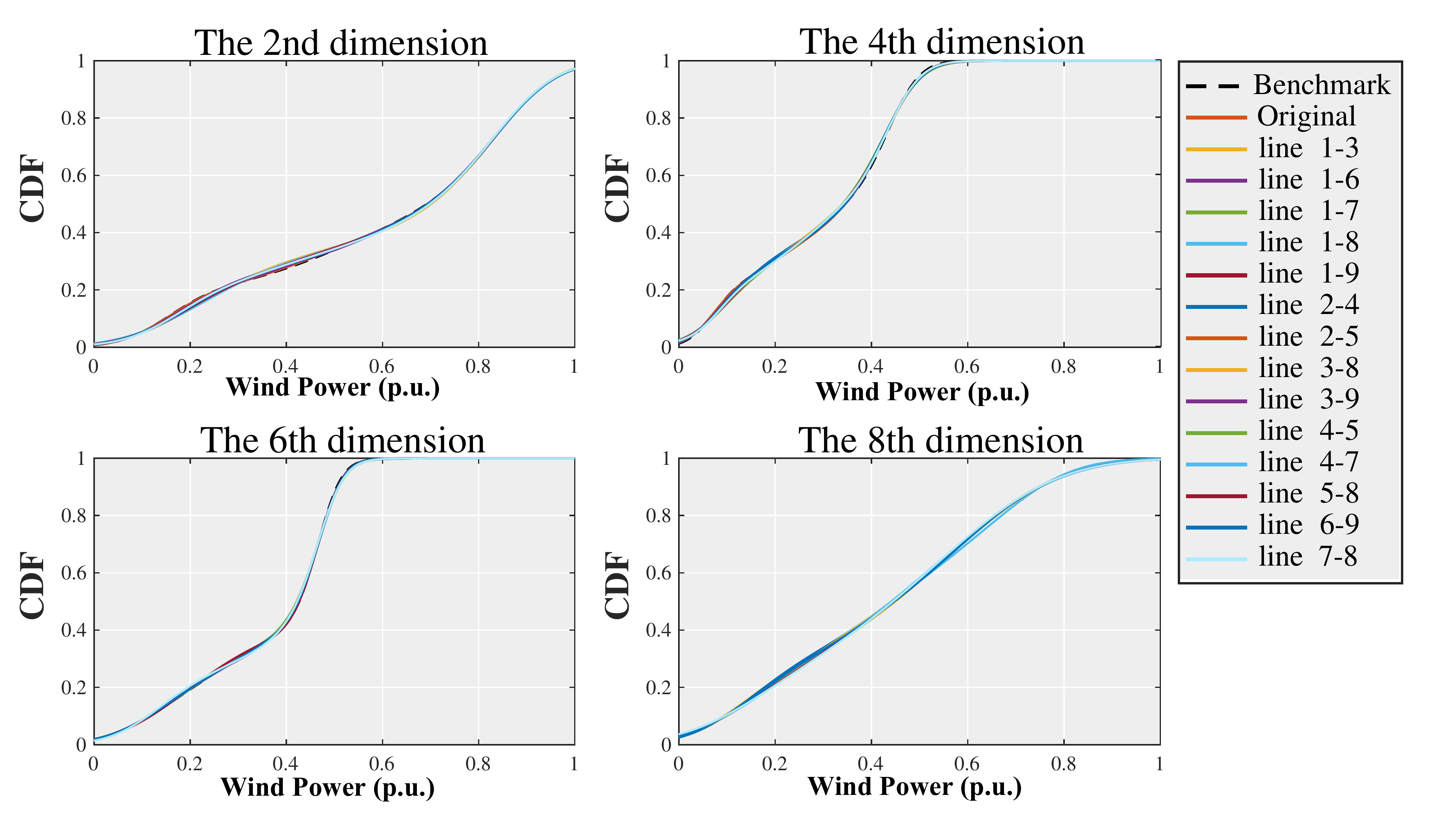}
\caption{The CDFs built after communication failures}
\label{case_failure}
\end{figure}

To verify the robustness of the proposed PPD EM algorithm, we first cut off communication lines to simulate communication failures. Then, we inspect the variations of the CDFs built by the proposed algorithm after the failures. Since the consensus of the proposed algorithm has already been verified, we still use the estimation result of the 1st WF as the representative. The CDFs built by the 1st WF after the communication failures are shown in Fig. \ref{case_failure}. In this figure, the benchmarks are the CDFs built by the centralized EM algorithm. Besides, legend `\textbf{\textit{Original}}' represents the CDFs built by the 1st WF when no failure occurs, and legend `line \textit{m}--\textit{i}' represents the CDFs built by the 1st WF when the communication between the \textit{m}-th WF and the \textit{i}-th WF fails. For example, legend `line 1-3' means that the communication between the 1st WF and the 3rd WF fails while other communication lines operate normally. As we can see, in Fig. \ref{case_failure}, the CDFs built by the 1st WF under different communication failures still coincide with the corresponding benchmarks and original CDFs. This proves that the proposed PPD EM algorithm can still maintain high accuracy after communication failures. Therefore, the robustness of the proposed algorithm is verified.

\vspace{-0.3cm}
\subsection{Verification of the Conditional Distribution}

Based on the established joint distribution via the proposed PPD EM algorithm, we derive each WF's conditional distribution of forecast error in (\ref{Conditional GMM}) by the PPD derivation algorithm proposed in \cite{Jia2019}. 
\vspace{-0.3cm}
\begin{figure}[h]
\centering 
\includegraphics[width=2.16in,center]{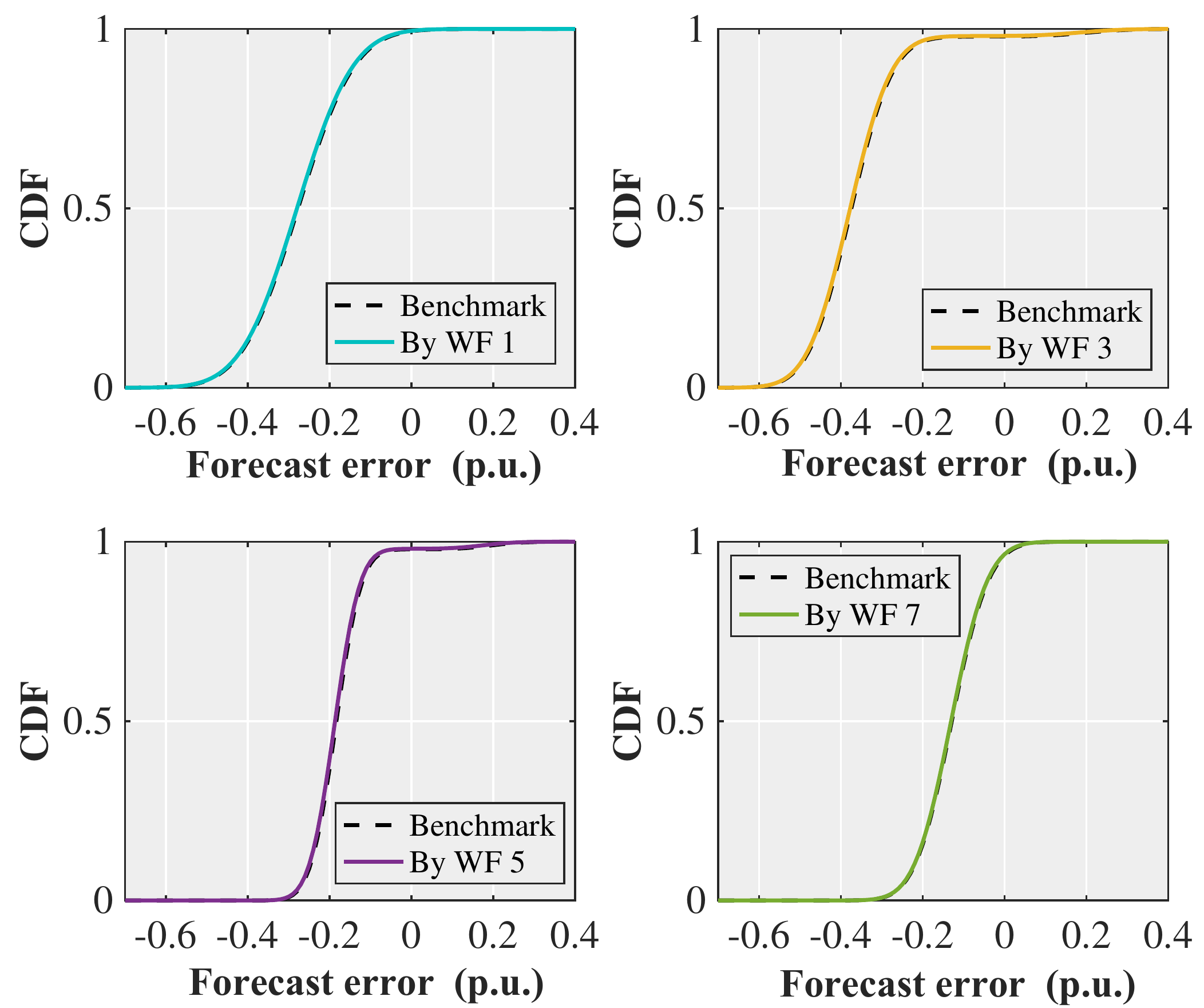}
\caption{The conditional CDF comparison between the centralized manner and the proposed PPD manner}
\label{case_conditional}
\end{figure}
The PPD derivation algorithm can enable each WF to obtain its conditional distribution from the joint one in the PPD manner. Here we illustrate the conditional CDF obtained by the 1st, 3rd, 5th, and 7th WFs in Fig. \ref{case_conditional}, where the benchmarks are all built in a centralized manner. The matches between the benchmarks and the CDFs built by WFs show that the joint distribution obtained via the proposed PPD EM algorithm is correct, so the conditional distributions derived from the joint distribution are correct as well.

\section{Conclusion}

Under the consideration of wind power correlation, estimating the conditional probability distribution of WF's forecast error requires the historical wind power and forecast data of all WFs. However, for the multi-correlated WFs spreading over a multi-regional interconnected grid, data barriers among the WFs belonging to different regions may exist. Therefore, we propose a PPD EM algorithm to estimate the joint probability distribution of the vertically partitioned wind power and forecast data. Then, we derive each WF's conditional probability distribution of forecast error from the joint one. To achieve this, we first formulate a distributed framework for the EM algorithm by reformulating the algorithm into local and global parts. Hereafter, the keys to developing a PPD EM algorithm are pointed out: calculating summations and inner products among the statistics of the correlated WFs in a PPD manner. After that, we design a PPD summation algorithm based on the additive homomorphic encryption and the average consensus algorithm. Thereafter, we develop a PPD inner product algorithm by leveraging the randomized binary hash mapping function and the average consensus algorithm. Combining the PPD summation and inner product algorithms, we finally propose the PPD EM algorithm. This algorithm can enable each WF to estimate the joint probability distribution of the wind power and forecast data of all the WFs in a fully distributed and privacy-preserving manner.

Compared with the centralized EM algorithm, the proposed algorithm not only has high accuracy but also is fully distributed because it only needs local communication between neighboring WFs. Moreover, it strictly protects the data privacy of every WF during communications. Furthermore, its robustness to communication failure is guaranteed by the consensus feature.

\ifCLASSOPTIONcaptionsoff
  \newind powerage
\fi

% trigger a \newind powerage just before the given reference


\bibliographystyle{IEEEtran}
\bibliography{IEEEabrv,paper}

\begin{thebibliography}{10}
\providecommand{\url}[1]{#1}
\csname url@samestyle\endcsname
\providecommand{\newblock}{\relax}
\providecommand{\bibinfo}[2]{#2}
\providecommand{\BIBentrySTDinterwordspacing}{\spaceskip=0pt\relax}
\providecommand{\BIBentryALTinterwordstretchfactor}{4}
\providecommand{\BIBentryALTinterwordspacing}{\spaceskip=\fontdimen2\font plus
\BIBentryALTinterwordstretchfactor\fontdimen3\font minus
  \fontdimen4\font\relax}
\providecommand{\BIBforeignlanguage}[2]{{%
\expandafter\ifx\csname l@#1\endcsname\relax
\typeout{** WARNING: IEEEtran.bst: No hyphenation pattern has been}%
\typeout{** loaded for the language `#1'. Using the pattern for}%
\typeout{** the default language instead.}%
\else
\language=\csname l@#1\endcsname
\fi
#2}}
\providecommand{\BIBdecl}{\relax}
\BIBdecl

\bibitem{WANG2018945}
X.~Wang, J.~Chang, X.~Meng, and Y.~Wang, ``Short-term
  hydro-thermal-wind-photovoltaic complementary operation of interconnected
  power systems,'' \emph{Appl. Energy.}, vol. 229, pp. 945 -- 962, 2018.

\bibitem{4530750}
H.~{Bludszuweit}, J.~A. {Dominguez-Navarro}, and A.~{Llombart}, ``Statistical
  analysis of wind power forecast error,'' \emph{IEEE Trans. Power Syst.},
  vol.~23, no.~3, pp. 983--991, Aug 2008.

\bibitem{1490597}
A.~Fabbri, T.~G.~S. Roman, J.~R. Abbad, and V.~H.~M. Quezada, ``Assessment of
  the cost associated with wind generation prediction errors in a liberalized
  electricity market,'' \emph{IEEE Trans. Power Syst.}, vol.~20, no.~3, pp.
  1440--1446, Aug 2005.

\bibitem{7862254}
Z.~Wang, C.~Shen, F.~Liu, X.~Wu, C.~Liu, and F.~Gao, ``Chance-constrained
  economic dispatch with non-gaussian correlated wind power uncertainty,''
  \emph{IEEE Trans. Power Syst.}, vol.~32, no.~6, pp. 4880--4893, Nov 2017.

\bibitem{7384775}
Z.~Wu, P.~Zeng, X.~Zhang, and Q.~Zhou, ``A solution to the chance-constrained
  two-stage stochastic program for unit commitment with wind energy
  integration,'' \emph{IEEE Trans. Power Syst.}, vol.~31, no.~6, pp.
  4185--4196, Nov 2016.

\bibitem{8315454}
Z.~Wang, C.~Shen, F.~Liu, J.~Wang, and X.~Wu, ``An adjustable
  chance-constrained approach for flexible ramping capacity allocation,''
  \emph{IEEE Trans. Sustain. Energy.}, vol.~9, no.~4, pp. 1798--1811, Oct 2018.

\bibitem{wang2018}
Z.~Wang, C.~Shen, and F.~Liu, ``A conditional model of wind power forecast
  errors and its application in scenario generation,'' \emph{Appl. Energy.},
  vol. 212, pp. 771--785, 2018.

\bibitem{8481551}
M.~Jia, C.~Shen, and Z.~Wang, ``A distributed probabilistic modeling algorithm
  for the aggregated power forecast error of multiple newly built wind farms,''
  \emph{IEEE Trans. Sustain. Energy.}, pp. 1--10, 2018.

\bibitem{SUN2019113842}
M.~Sun, C.~Feng, and J.~Zhang, ``Conditional aggregated probabilistic wind
  power forecasting based on spatio-temporal correlation,'' \emph{Appl.
  Energy.}, vol. 256, p. 113842, 2019.

\bibitem{ZHANG2019229}
J.~Zhang, J.~Yan, D.~Infield, Y.~Liu, and F.~sang Lien, ``Short-term
  forecasting and uncertainty analysis of wind turbine power based on long
  short-term memory network and gaussian mixture model,'' \emph{Appl. Energy.},
  vol. 241, pp. 229 -- 244, 2019.

\bibitem{982193}
J.~{Contreras}, A.~{Losi}, M.~{Russo}, and F.~F. {Wu}, ``Simulation and
  evaluation of optimization problem solutions in distributed energy management
  systems,'' \emph{IEEE Trans. Power Syst.}, vol.~17, no.~1, pp. 57--62, Feb
  2002.

\bibitem{6939252}
A.~{Ahmadi-Khatir}, A.~{Conejo}, and R.~{Cherkaoui}, ``Multi-area unit
  scheduling and reserve allocation under wind power uncertainty,'' in
  \emph{2014 IEEE PES General Meeting | Conference Exposition}, July 2014, pp.
  1--1.

\bibitem{7990367}
Z.~{Wang}, F.~{Liu}, S.~H. {Low}, C.~{Zhao}, and S.~{Mei}, ``Distributed
  frequency control with operational constraints, part ii: Network power
  balance,'' \emph{IEEE Trans. Smart Grid.}, vol.~10, no.~1, pp. 53--64, Jan
  2019.

\bibitem{RAHMAN201720}
M.~Rahman and A.~Oo, ``Distributed multi-agent based coordinated power
  management and control strategy for microgrids with distributed energy
  resources,'' \emph{Energy Conversion and Management}, vol. 139, pp. 20 -- 32,
  2017.

\bibitem{chen2019distributed}
H.~Chen, X.~Wang, Z.~Li, W.~Chen, and Y.~Cai, ``Distributed sensing and
  cooperative estimation/detection of ubiquitous power internet of things,''
  \emph{Protection and Control of Modern Power Systems}, vol.~4, no.~1, p.~13,
  2019.

\bibitem{Clifton2002}
C.~Clifton, M.~Kantarcioglu, J.~Vaidya, X.~Lin, and M.~Y. Zhu, ``Tools for
  privacy preserving distributed data mining,'' \emph{SIGKDD Explor. Newsl.},
  vol.~4, no.~2, pp. 28--34, Dec. 2002.

\bibitem{Lin2005}
X.~Lin, C.~Clifton, and M.~Y. Zhu, ``Privacy-preserving clustering with
  distributed em mixture modeling,'' \emph{Knowledge and Information Systems},
  vol.~8, no.~1, pp. 68--81, Jul 2005.

\bibitem{8029559}
K.~L. Leemaqz, S.~X. Lee, and G.~J. McLachlan, ``Corruption-resistant privacy
  preserving distributed em algorithm for model-based clustering,'' in
  \emph{2017 IEEE Trustcom/BigDataSE/ICESS}, Aug 2017, pp. 1082--1089.

\bibitem{Yang2012}
B.~Yang, I.Sato, and H.Nakagawa, ``Privacy-preserving em algorithm for
  clustering on social network,'' in \emph{Advances in Knowledge Discovery and
  Data Mining}.\hskip 1em plus 0.5em minus 0.4em\relax Berlin, Heidelberg:
  Springer Berlin Heidelberg, 2012, pp. 542--553.

\bibitem{5410606}
Y.~Weng, L.~Xie, and W.~Xiao, ``Diffusion scheme of distributed em algorithm
  for gaussian mixtures over random networks,'' in \emph{2009 IEEE
  International Conference on Control and Automation}, Dec 2009, pp.
  1529--1534.

\bibitem{Weng2011}
Y.~Weng, W.~Xiao, and L.~Xie, ``Diffusion-based em algorithm for distributed
  estimation of gaussian mixtures in wireless sensor networks,''
  \emph{Sensors}, vol.~11, no.~6, pp. 6297--6316, 2011.

\bibitem{5606758}
S.~S. Pereira, S.~Barbarossa, and A.~Pagés-Zamora, ``Consensus for distributed
  em-based clustering in wsns,'' in \emph{2010 IEEE Sensor Array and
  Multichannel Signal Processing Workshop}, Oct 2010, pp. 45--48.

\bibitem{4558075}
D.~Gu, ``Distributed em algorithm for gaussian mixtures in sensor networks,''
  \emph{IEEE Trans. Neural Netw.}, vol.~19, no.~7, pp. 1154--1166, July 2008.

\bibitem{Jia2019}
M.~Jia, C.~Shen, and Z.~Wang, ``A distributed privacy-preserving incremental
  update algorithm for probability distribution of wind power forecast error,''
  \emph{[Online]}, 2019, Available: https://arxiv.org/abs/1905.06420.

\bibitem{techrep2}
J.~Bilmes, ``A gentle tutorial on the em algorithm and its application to
  parameter estimation for gaussian mixture and hidden markov models,''
  University of Berkeley, Tech. Rep. ICSI-TR-97-02, 1997.

\bibitem{4118472}
R.~Olfati-Saber, J.~A. Fax, and R.~M. Murray, ``Consensus and cooperation in
  networked multi-agent systems,'' \emph{Proceedings of the IEEE}, vol.~95,
  no.~1, pp. 215--233, Jan 2007.

\bibitem{5601452}
F.~Xu, S.~Zeng, S.~Luo, C.~Wang, Y.~Xin, and Y.~Guo, ``Research on secure
  scalar product protocol and its' application,'' in \emph{2010 6th
  International Conference on Wireless Communications Networking and Mobile
  Computing (WiCOM)}, Sept 2010, pp. 1--4.

\bibitem{Paillier1999}
P.~Paillier, ``Public-key cryptosystems based on composite degree residuosity
  classes,'' in \emph{Advances in Cryptology --- EUROCRYPT '99}, J.~Stern,
  Ed.\hskip 1em plus 0.5em minus 0.4em\relax Berlin, Heidelberg: Springer
  Berlin Heidelberg, 1999, pp. 223--238.

\bibitem{XIAO200733}
L.~Xiao, S.~Boyd, and S.-J. Kim, ``Distributed average consensus with
  least-mean-square deviation,'' \emph{Journal of Parallel and Distributed
  Computing}, vol.~67, no.~1, pp. 33 -- 46, 2007.

\bibitem{Goemans1995}
M.~X. Goemans and D.~P. Williamson, ``Improved approximation algorithms for
  maximum cut and satisfiability problems using semidefinite programming,''
  \emph{J. ACM}, vol.~42, no.~6, pp. 1115--1145, Nov. 1995.

\bibitem{MARUKATAT2013}
M.~Sanparith and M.~Ithipan, ``Fast nearest neighbor retrieval using randomized
  binary codes and approximate euclidean distance,'' \emph{Pattern Recognition
  Letters}, vol.~34, no.~9, pp. 1101 -- 1107, 2013.

\bibitem{Clifton20022}
J.~Vaidya and C.~Clifton, ``Privacy preserving association rule mining in
  vertically partitioned data,'' in \emph{Proceedings of the Eighth ACM SIGKDD
  International Conference on Knowledge Discovery and Data Mining}, ser. KDD
  '02.\hskip 1em plus 0.5em minus 0.4em\relax New York, NY, USA: ACM, 2002, pp.
  639--644.

\bibitem{4542554}
T.~C. {Aysal}, M.~J. {Coates}, and M.~G. {Rabbat}, ``Distributed average
  consensus with dithered quantization,'' \emph{IEEE Transactions on Signal
  Processing}, vol.~56, no.~10, pp. 4905--4918, Oct 2008.

\end{thebibliography}
\end{document}